\documentclass[conference]{IEEEtran}
\IEEEoverridecommandlockouts
\IEEEpubid{\makebox[\columnwidth]{979-8-3503-2445-7/23/\$31.00~\copyright2023 IEEE \hfill} \hspace{\columnsep}\makebox[\columnwidth]{ }}

\usepackage{url}
\newcommand\R{{\mathbb R}}

\newcommand\N{{\mathbb N}}
\newcommand\E{{\mathbb E}}

\usepackage{algorithm}
\usepackage{algorithmic}

\usepackage{newfloat}
\usepackage{listings}
\usepackage{tikz}

\makeatletter
\newcommand\footnoteref[1]{\protected@xdef\@thefnmark{\ref{#1}}\@footnotemark}
\makeatother

\usepackage{wrapfig}
\usepackage{varwidth}
\usepackage{graphicx}
\usepackage{booktabs}
\usepackage{amsmath}
\usepackage{amsthm}
\usepackage{amsfonts}
\usepackage{epsfig}
\usepackage{graphicx}
\usepackage{subfigure}
\usepackage[bottom]{footmisc}
\usepackage{balance}

\usepackage{multirow}
\usepackage{xspace}
\usepackage[singlelinecheck=off]{caption}
\DeclareCaptionType{copyrightbox}
\usepackage{pifont}
 \newtheorem{problem}{Problem}

\newcommand{\spara}[1]{\smallskip\noindent{\bf #1}}

\newcommand{\squishlist}{
 \begin{list}{$\bullet$}
  {  \setlength{\itemsep}{0pt}
     \setlength{\parsep}{3pt}
     \setlength{\topsep}{3pt}
     \setlength{\partopsep}{0pt}
     \setlength{\leftmargin}{2em}
     \setlength{\labelwidth}{1.5em}
     \setlength{\labelsep}{0.5em}
} }
\newcommand{\squishlisttight}{
 \begin{list}{$\bullet$}
  { \setlength{\itemsep}{0pt}
    \setlength{\parsep}{0pt}
    \setlength{\topsep}{0pt}
    \setlength{\partopsep}{0pt}
    \setlength{\leftmargin}{2em}
    \setlength{\labelwidth}{1.5em}
    \setlength{\labelsep}{0.5em}
} }

\newcommand{\squishdesc}{
 \begin{list}{}
  {  \setlength{\itemsep}{0pt}
     \setlength{\parsep}{3pt}
     \setlength{\topsep}{3pt}
     \setlength{\partopsep}{0pt}
     \setlength{\leftmargin}{1em}
     \setlength{\labelwidth}{1.5em}
     \setlength{\labelsep}{0.5em}
} }

\newcommand{\squishend}{
  \end{list}
}

\newcommand{\eat}[1]{}

\newcounter{ccc}

\newcommand{\bigO}{\mathcal{O}}

\begin{document}\sloppy

\title{Tripletformer for Probabilistic Interpolation of Irregularly sampled Time Series}
\author{ Vijaya Krishna Yalavarthi, Johannes Burchert, Lars Schmidt-Thieme}

\author{\IEEEauthorblockN{Vijaya Krishna Yalavarthi}
	\IEEEauthorblockA{\textit{ISMLL, University of Hildesheim} \\
	Germany \\
yalavarthi@ismll.uni-hildesheim.de}
	\and
	\IEEEauthorblockN{Johannes Burchert}
	\IEEEauthorblockA{\textit{ISMLL, University of Hildesheim} \\
		Germany \\
		burchert@ismll.uni-hildesheim.de}
	\and
	\IEEEauthorblockN{Lars Schmidt-Thieme}
	\IEEEauthorblockA{\textit{ISMLL, University of Hildesheim} \\
		Germany \\
		schimdt-thieme@ismll.uni-hildesheim.de}
}

\maketitle
\IEEEpubidadjcol
\begin{abstract}

Irregularly sampled time series data with missing values is a observed in many fields like healthcare, astronomy, and climate science.
Interpolation of these types of time series is crucial for tasks such as root cause analysis and medical diagnosis, as well as for smoothing out irregular or noisy data.
To address this challenge, we present a novel encoder-decoder architecture called ``Tripletformer'' for probabilistic interpolation of irregularly sampled time series with missing values. 
This attention-based model operates on sets of observations, where each element is composed of a triple of time, channel, and value.
The encoder and decoder of the Tripletformer are designed with attention layers and fully connected layers, enabling the model to effectively process the presented set elements.
We evaluate the Tripletformer against a range of baselines on multiple real-world and synthetic datasets and show that it produces more accurate and certain interpolations.
Results indicate an improvement in negative loglikelihood error by up to $32\%$ on real-world datasets and $85\%$ on synthetic datasets when using the Tripletformer compared to the next best model.
\end{abstract}


\section{Introduction}
\label{sec:intro}

In domains such as medical applications~\cite{YS18}, multivariate time series (MTS) are frequently observed with irregularity.
This implies that the variables (or sensors) within the time series are observed at irregular time intervals and often contain missing values during the alignment process.
We refer to such time series as Irregularly Sampled Time Series with Missing Values (IMTS).
Practitioners aim to identify data that was not collected in the initial phase of data collection for better analysis.
As an example, blood glucose levels are typically measured multiple times a day using a glucose meter, but due to a patient's forgetfulness or other reasons, there may be missing data points. The healthcare provider may want to determine the missing glucose level values in order to better understand the patient's glucose control and make more informed treatment decisions. Probabilistic interpolation models can be used to estimate the probable values of the missing glucose level measurements, along with the uncertainty surrounding the predictions allowing the healthcare provider to avoid overly confident predictions.
Furthermore, considering that sensor observations inherently involve measurement errors, it becomes essential to quantify the uncertainty associated with interpolated predictions. This is particularly important in diverse fields such as engineering and environmental applications~\cite{LA17}.

\begin{figure}[t]
	\centering
	\includegraphics[clip, width=\columnwidth]{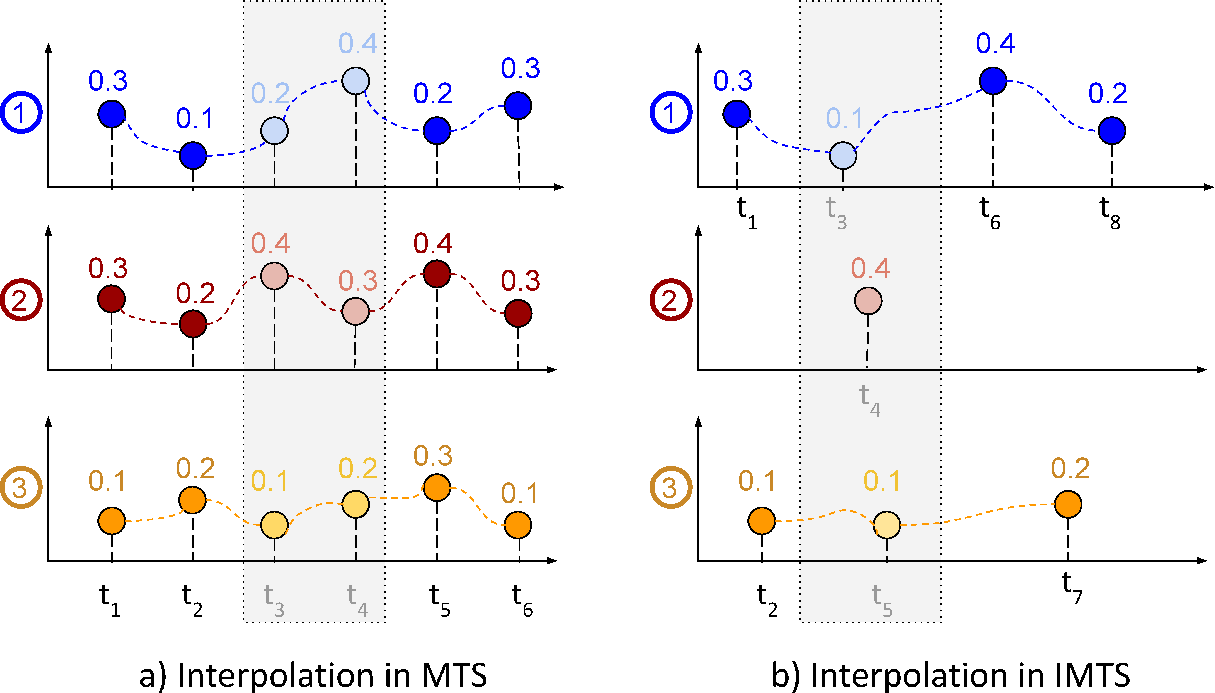}
	\caption{Interpolation in Multivariate Time Series (a) and Irregularly sampled Time Series (b). In (a) all the channels in the time series are observed at times $t_1, t_2, t_5, t_6$ and we need to interpolate the values for all the channels at $t_3, t_4$. Where as in (b), channel 1 is observed at times $t_1, t_6, t_8$, and channel 3 is observed at $t_2, t_7$. We did not make any observation in channel 2. However, we need to interpolate the values for channels 1,2 and 3 at time points $t_3$, $t_4$, and $t_5$ respectively.}
	\label{fig:motivation}
\end{figure}

Deep learning models have been widely studied for imputation tasks in general multivariate time series (MTS) with missing values, where the time series is observed at regular intervals, the sparsity (number of missing values over the number of observed values) is small, and all the channels are of equal lengths. However, these models are not well suited for irregularly sampled time series with missing values (IMTS) due to their unique properties. IMTS often consist of non-periodic observations in a channel, extremely sparse channels when trying to align the IMTS, and variable channel lengths in a single example. These properties make modeling IMTS difficult for standard deep learning models.
Figure~\ref{fig:motivation} illustrates the differences between interpolation in IMTS and MTS. 

In the domain of IMTS, most research has focused on classification techniques~\cite{SM21,LM16,RC19,KM20,HM20,SM19,VJ22}. However, some studies have delved into the area of deterministic interpolation~\cite{SM21,CR18,RC19}, but a few on the probabilistic interpolation. Gaussian Process Regression models~\cite{BC07,WR06} were used traditionally for the task~\cite{DP14}. Tashiro et. al.\cite{TS21} proposed a score based diffusion model, CSDI, however it only produced homoscedastic variance meaning it provides same uncertainty for all the outputs. Recently, Sukla et. al.~\cite{SM22} introduced the Heteroscedastic Variational Autoencoder (HETVAE) which uses the Uncertainty aware multi Time Attention Network (UnTAN) to produce heteroscedastic variance for probabilistic interpolation in IMTS. HETVAE struggles to effectively learn cross-channel interactions because its core component UnTAN is built upon existing multi-Time Attention Network (mTAN)~\cite{SM21} which employs a separate time attention model for each channel. Through our experiments, we have found that encoding observations from all channels using a single encoder leads to more accurate interpolations.

In this work, we propose a novel encoder-decoder architecture, that we call Tripletformer for the probabilistic interpolation of IMTS . As the name suggests, our Tripletformer operates on the observations that are in triplet form (time, channel and value of the observation).
We model time and channel as the independent variables, and the observation measurement (value) as the dependent variable.
The encoder of the Tripletformer learns the interactions among the inter and intra channel observations of the IMTS simultaneously, while the decoder produces the probability distributions of the dependent variable over the set of reference (or target) independent variables. We overcome the computational complexity bottleneck of canonical attention by using induced multi-head attention block~\cite{LL19} that has near linear computational complexity (that we explain in Section~\ref{sec:imab}).

We evaluate the proposed Tripletformer over multiple $3$ real-world and $2$ synthetic datasets at varying observation levels, and two different missing patterns: random missing and burst missing. The Tripletformer is compared with the state-of-the-art probabilistic interpolation models, HETVAE and CSDI, and a range of baselines using Negative Loglikelihood loss (NLL) and CRPS as the evaluation metric. Our experimental results attest that the proposed Tripletformer provides significantly better interpolations compared to its competitors.
Our contributions are summarized as follows:
\begin{itemize}
	\item We propose a novel model called Tripletformer that can learn both inter and intra channel interactions simultaneously in the IMTS by operating on the observations for probabilistic interpolation.
	
	\item To overcome the computational complexity of multi-head attention, we employ an induced multi-head attention mechanism~\cite{LL19}. This improves predictions while efficiently addressing the computational bottleneck.
	
	\item We perform extensive experimental evaluation on $3$ real and $2$ synthetic IMTS datasets, under two different missing patterns: random and burst missing. Our results attest that Tripletformer provides up to $32\%$ and $85\%$ improvement in NLL over real and synthetic IMTS datasets, respectively compared to the next best model HETVAE.
\end{itemize}
We provide the implementation of Tripletformer in \\\url{https://github.com/yalavarthivk/tripletformer}.
\section{Related Work}
\label{sec:related}

The focus of this work is on probabilistic interpolation of irregularly sampled time series with missing values (IMTS). Majority of the recent works in this field have mainly focused on deterministic interpolation techniques for IMTS.

For example, \cite{YZ18} and~\cite{CW18} applied Multidirectional Recurrent Neural Network and Bidirectional Recurrent Neural Networks for IMTS imputation by using a GRU decay (GRU-D)~\cite{CP18} model as the underlying architecture to handle the irregularity in the series.~\cite{SM19} proposed Interpolation-Prediction networks, which consist of a semi-parametric model to share information across multiple channels of an IMTS. This approach is similar to a multivariate Gaussian process~\cite{BC07}, but does not use a positive definite co-variance matrix.

These methods provide deterministic interpolation, which means they cannot be used to find the uncertainty around the predictions, as done in probabilistic interpolation. This is a limitation of these models, as probabilistic interpolation provides more information about the confidence of the predictions, which can be useful in many real-world applications. The current work aims to propose a novel probabilistic interpolation method for IMTS.

Researchers have proposed various models for probabilistic interpolation of time series data that utilize Variational Autoencoders (VAEs) and can produce homoscedastic variance, which means the uncertainty in the predictions is constant across all predictions.
In 2018,~\cite{CR18} proposed using a Neural Ordinary Differential Equations (ODE) network for modeling time series by using a continuous time function in the hidden state. Later, in 2019,~\cite{RC19} proposed an architecture with ODE-RNN as the encoder which uses Neural ODE to model the latent state dynamics and an RNN to update it when a new observation is made. In 2020,~\cite{LW20} proposed using stochastic differential equations (SDEs) which generalize ODEs and are defined using non-standard integrals, usually relying on It{$\hat{\textnormal{o}}$} calculus. In 2021,~\cite{NB21} proposed Neural ODE processes (NDPs) by combining Neural ODEs with Neural Processes~\cite{GS18}. NDPs are a class of stochastic processes that are determined by a distribution over Neural ODEs. Other than ODE based models,~\cite{SM21} proposed Multi-Time Attention Networks (mTAN), which uses a temporal attention network at both encoder and decoder to produce deterministic interpolations. Recently in 2021,~\cite{TS21} a score based diffusion model called CSDI was proposed. The main idea behind this model is to use a score function to evaluate the likelihood of different imputations and to select the one that is most accurate.

While VAE and diffusion-based models can produce homoscedastic variance, it is also possible to produce heteroscedastic variance, where the uncertainty in the predictions varies across different predictions.
One approach to producing heteroscedastic variance is to output the distribution parameters, rather than a fixed point estimate. Gaussian Process Regression (GPR)~\cite{WR06} provides full joint posterior distributions over interpolation outputs, which can be used to produce heteroscedastic variance. GPR models can be directly implemented on IMTS by applying one model for each channel. Multi-task Gaussian Process~\cite{BC07} have been studied for the multivariate setting but they require a positive definite covariance matrix. In~\cite{DP14}, D{\"u}richen et.al., studied various kernels for the Multi-task Gaussian Process for the analysis of IMTS.
Finally, HETVAE model, proposed in~\cite{SM22} is the first work that deals with the problem of probabilistic interpolation in IMTS and provides heteroscedastic variance. The HETVAE model consists of an Uncertainty-aware multi-Time Attention Networks (UnTAN) that uses a combination of probabilistic and deterministic paths to output heteroscedastic variance.

In this work, an attention-based model called Tripletformer is proposed for probabilistic interpolation of irregularly sampled time series with missing values (IMTS). Tripletformer operates on the set of observations and produces outputs with heteroscedastic variance.

\section{Background and Preliminaries}
\label{sec:prilim}

\subsection{IMTS as set of observations}

\begin{figure}[t]
	\centering
	\includegraphics[clip, width=0.9\columnwidth]{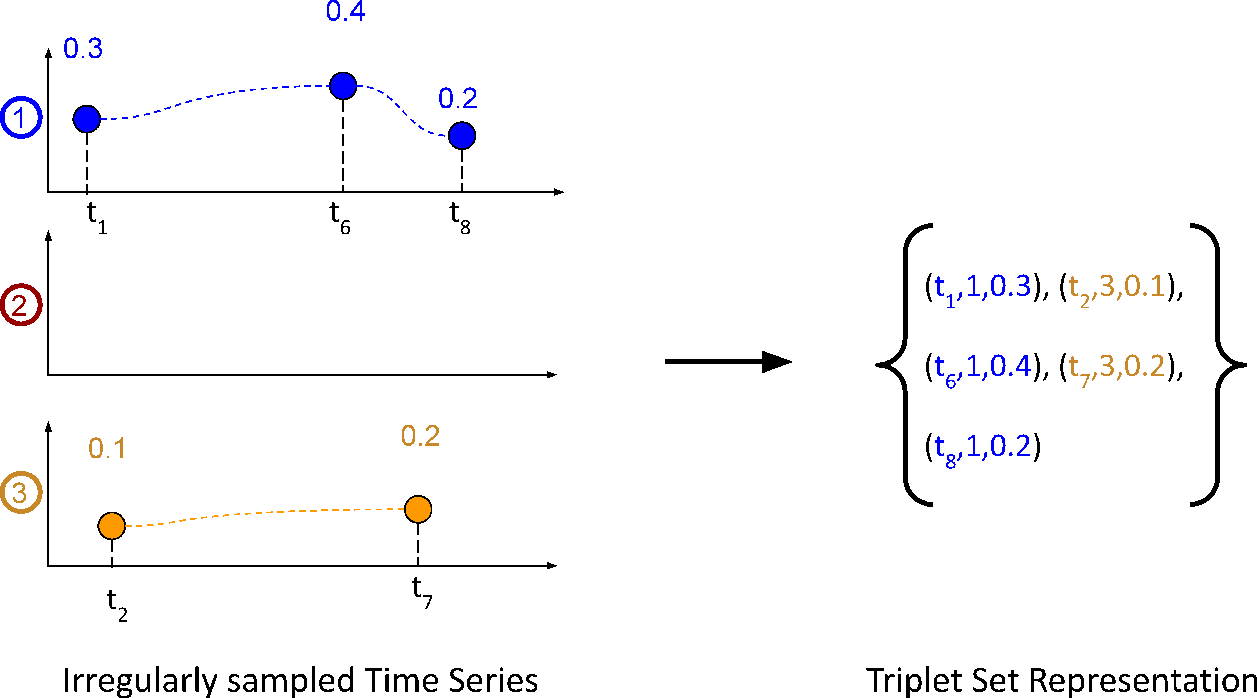}
	\caption{Demonstration of IMTS as set of observations.}
	\label{fig:set_rep}
\end{figure}

Our set representation of IMTS follows~\cite{HM20}. Given an IMTS dataset $\mathcal{X} = \{X^n\}_{n=1}^{N}$ of $N$ many examples, an IMTS $X^n$ can be formulated as set of observations $x^n_i$ such that $X^n = \{x^n_1, x^n_2,...x^n_{|X^n|}\}$. We omit superscript $n$ when the context is clear.
Each observation $x_i$ is a triple consisting of time, channel and value (observation measurement), $x_i = (t_i, c_i, u_i)$, with $t_i \in \mathcal{T} = \{\tau_1, \tau_2, ..., \tau_{|\mathcal{T}|}\}$, $\mathcal{T}$ being the set of all the observation times in the series $X$, $c_i \in \mathcal{C} = \{1,2,3,...,C\}$, $\mathcal{C}$ being the set of channels in all the IMTS dataset $\mathcal{X}$. In Figure~\ref{fig:set_rep}, we demonstrate the representation of IMTS as set of observations.

\subsection{Probabilistic Interpolation of IMTS}

In deterministic interpolation, one wants to estimate the most probable value of the dependent variable. While deterministic interpolation provides a single estimate, probabilistic interpolation outputs the probability distribution that explains the prediction's uncertainty.

Given an Irregularly sampled time series $X \in \mathcal{X}$, independent variables ($w = (t',c')$) in the target observations that are time $t' \notin \mathcal{T}$, $\min(\mathcal{T})<t' < \max(\mathcal{T})$, channel indicator $c' \in \mathcal{C}$; the goal is to predict the probability distribution of the dependent variable $u'$ i.e., find $\hat{Pr}(u'|X,w)$.

\begin{problem}[Probabilistic Interpolation in Time Series]
	Given i) a dataset $\mathcal{D}^{train} \sim \rho$ drawn from an unknown distribution $\rho$, an instance $(\{x\}^*,\{x'\}^*) \in \mathcal{D}^{train}$ with $x, x' \in \Omega$, $x = (t,c,u), x' = (t',c',u')$, $\Omega = \R_+\times\N\times\R$; $t, t' \in [\R_+, \R_+]$, $t \ne t'$,
	ii) a loss function $\mathcal{L} : \R\times \mathcal{F} \rightarrow \R$ example negative loglikelihood loss, $\mathcal{F}$ denotes a probability distribution. Find a model $f: \Omega^*\times\R^*_+\times\N^* \rightarrow \mathcal{F}$ such that the expected loss is minimized $\E_{(\{x\}^*, \{x'\}^*) \sim \rho} \quad \mathcal{L}(\{u'\}^*, f(\{x\}^*,\{(t', c')\}^*))$.
\end{problem}

\subsection{Multi-head Attention}

Multi-head attention (MHA)~\cite{VS17} jointly attends different positions from different representation spaces. MHA with $h$ many heads is defined as

\begin{equation}
\textnormal{MHA}(q, k, v) = \textnormal{Concat}(\textnormal{head}_1,..., \textnormal{head}_h)\theta^o
\end{equation}

where $q \in \R^{L_q \times d_q}$, $k \in \R^{L_k \times d_k}$, $v \in \R^{L_k\times d_v}$ are called query, key and value sequences with lengths of $L_q$, $L_k$, $L_k$, and embedding dimension of $d_q$, $d_k$, $d_v$ respectively. $\textnormal{head}_i, i\le h$ is the $i^{th}$ head which is defined as:

\begin{align}
\textnormal{head}_i & = \textnormal{Att}(q\theta_i^q, k\theta_i^k, v\theta_i^v) = Softmax({A})v\theta_i^v \\
\textnormal{Att} & = \frac{q\theta_i^q\left(k\theta_i^k\right)^T}{\sqrt{d}}
\end{align}

where $\theta_i^q \in \R^{d_q\times d}$, $\theta_i^k \in \R^{d_k\times d}$, $\theta_i^v \in \R^{d_v\times d}$ and $\theta^o \in \R^{d\times d_o}$ are the learned parameters, $d$ is the hidden dimension of the projection spaces, and $d_o$ is the dimension of the output sequence after MHA.

Attention matrix ($A$) which requires multiplication of $L_q\times d$ and $L_k\times d$ sized matrices has a computational complexity of $\bigO(L_qL_kd)$. When both $L_q$ and $L_k$ are arbitrarily large, computing $A$ will be a bottleneck.

%
%

\subsection{Multihead Attention Block (MAB)}
MAB~\cite{VS17,LL19} consists of two sublayers: i) Multihead Attention (MHA)~\cite{VS17},  and ii) a pointwise feed forward layer (MLP) as shown in Figure~\ref{fig:fact_attn_demo}(a). We have a residual connection around both the sublayers.

\begin{align}
	\textnormal{MAB}(q, k, v) &= \alpha(H + \textnormal{MLP}(H)) \nonumber \\
	\textnormal{where} \quad H &= \alpha(q + \textnormal{MHA}(q, k, v))
\end{align}

with $q,k,v$ being query, key and value sequences and, $\alpha$ is a non-linear activation function. 

In MHA, the attention matrix ($A$) which requires the multiplication of $L_q\times d$ and $L_k\times d$ sized matrices has a computational complexity of $\bigO(L_qL_kd)$. $L_q$, $L_k$, $d$ are the query length, key length and embedding dimensions respectively. When both $L_q$ and $L_k$ are large, computing $A$ will be a bottleneck because of its quadratic computational complexity.

\subsection{Induced Multihead Attention Block (IMAB)}
\label{sec:imab}

Because computing $A$ in MHA is infeasible for long sequences, \cite{LL19} proposed the Induced Multihead Attention Block (IMAB) that consists of two MABs and $l$ many $d_h$ dimensional induced points $h \in \R^{l\times d_h}$ which are trainable parameters. In IMAB, the attention happens as follows: first induced points attend to the actual keys and values, later queries attend to the induced points. IMAB is given as:
\begin{align}
	\textnormal{IMAB}(q, k, v) & = \textnormal{MAB}(q, \mathcal{H}, \mathcal{H}) \nonumber \\
	\textnormal{where} \quad \mathcal{H} &= \textnormal{MAB}(h, k, v) \label{eq:imab}
\end{align}

\begin{figure}[t]
	\centering
	\includegraphics[clip, width=\columnwidth]{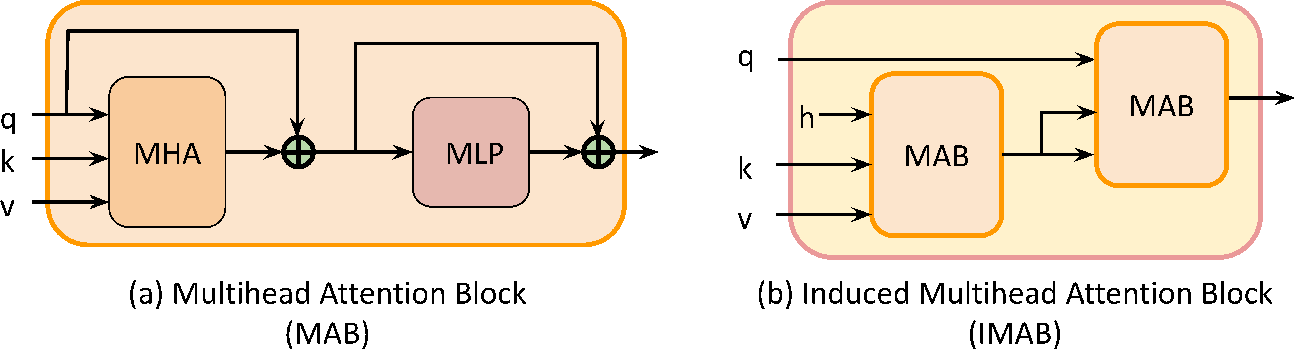}
	\caption{Architectures of Multihead Attention Block (a) and Induced Multihead Attention Block (b)~\cite{LL19}}
	\label{fig:fact_attn_demo}
\end{figure}

The architecture of IMAB is shown in Figure~\ref{fig:fact_attn_demo}(b). The computational complexity of the IMAB is $\bigO(L_qld + L_kld)$ which is near linear when $l \ll L_q, L_k$.

IMAB is the optimal choice of model because i) it reduces the computational complexity and ii) provides learned restricted attention phenomena. Probsparse attention~\cite{ZZ21} which provides restricted attention is the widely used model for the time series forecasting. It ranks the key tokens by their impact on the query, and restricts its attention to the top-$K$ tokens only rather than the entire key set. It was shown that using restricted attention provides similar or better results compared to the canonical attention (see~\cite{ZZ21} for more details). However, the main drawback of Probsparse attention is, it assumes the lengths of all the series in a batch are equal which is not the case while working with IMTS. In order to have a consistent batch, we pad the series of smaller lengths with zeros, and the sparse attentions treat observations and zero padding in the same manner for sampling the important key tokens which is incorrect.
On the other hand, when we are operating on sets, it is not trivial to find the key tokens that are important to the query, manually. Hence, one can learn a set of clusters called reference clusters ($\mathcal{H}$ in equation~\ref{eq:imab}) using $h$ that act as basis for the restriction. Then, the queries attend to those reference clusters rather than the entire key set.




\section{Proposed Model: Tripletformer}
\label{sec:prop_model}

Our proposed Tripletformer holds an encoder-decoder architecture for the problem of probabilistic interpolation in IMTS. The model utilizes the irregular sampling of observations by operating on the observations directly.

\begin{figure}[t]
	\centering
	
	\includegraphics[clip, width=0.8\columnwidth]{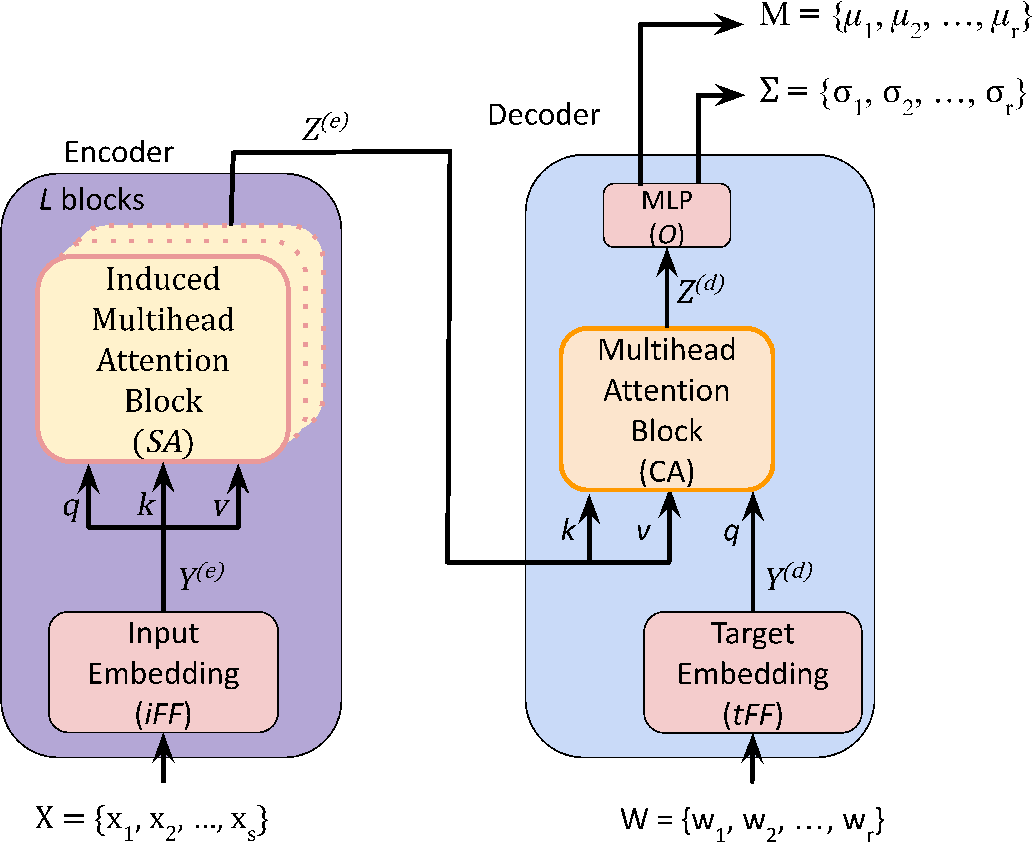}
	\caption{Tripletformer architecture. Encoder (left) takes the set of observations $X$, and output their embeddings $Z^{(e)}$. Decoder (right) takes $Z^{(e)}$, target queries ($W$), and produces the mean $\mathrm{M}$ and standard deviation $\Sigma$ corresponding to $W$.}
	\label{fig:Tripletformer}
\end{figure}

Our encoder $E$, similar to that of the transformer model~\cite{VS17}, maps the input time series $X$ which is a set of $s = |X|$ many observations $\{x_1, ..., x_s\}$ to a set of continuous representations $Z^{(e)} = \{z^{(e)}_1, ..., z^{(e)}_s\}$ corresponding to $X$. Our decoder outputs the parameters of the probability distribution of the target observation value $\hat{Pr}(u'|Z^{(e)},w)$ conditioning up on the encoder output $Z^{(e)}$, and independent variables $w = (t', c')$ of the target observation $x'$.
The architecture of the proposed model is shown in Figure~\ref{fig:Tripletformer}. The model components are explained in more detail in the following sections of the paper.

\subsection{Encoder}
\label{sec:prop_encoder}

Our encoder consists of i) an input feed forward embedding layer ($iFF$) and ii) a Self-attention layer ($SA$) which is an IMAB.


The $iFF$ in the encoder is a point-wise feed-forward layer that provides learned embeddings to the set elements. To be able to work with the $iFF$, each observation $x$ in the input time series $X$ is converted into a vector $\mathrm{x} \in \R^{C+2}$ by concatenating the time, channel indicator, and value of the observation. The channel indicator is one-hot encoded. The $iFF$ takes the set of vectors $\mathrm{X}$ as input and outputs its embeddings $Y^{(e)} = {y^{(e)}_1, ..., y^{(e)}_s}$. These input embeddings are then passed through a Self-Attention layer $SA$ which consists of $L$ Induced Multihead Attention Blocks. This produces the latent embeddings $Z^{(e)} = {z^{(e)}_1, ..., z^{(e)}_s}$.


Using an attention mechanism is a principled approach for the interpolation problem in IMTS. 
In IMTS, the length of a series (size of the set) varies, and a model that does not depend on the length of the series is needed in order to learn the latent embeddings. One could sort the set elements over time and apply a Convolutional Neural Network (CNN) or Recurrent Neural Network (RNN), but these methods are not useful for interpolation tasks because the query time point is in between the observation times where convolution (or RNN) is already applied. The attention mechanism, on the other hand, allows each observation to attend to all other observations in the set, making it an ideal choice for interpolation tasks. 

\subsection{Decoder}

The decoder takes the encoder output $Z^{(e)}$, and $r$ many target queries ${W = (w_1, ..., w_r)}$ as inputs. It then outputs the distribution parameters of the target values ${U' = u'_1, ..., u'_r}$. Specifically, the decoder outputs $\mathrm{M} = {\mu_1, ..., \mu_r}$, and $\Sigma = {\sigma_1, ..., \sigma_r}$ which are the corresponding mean and standard deviation assuming the underlying distribution is Gaussian. In the current work we assume that the underlying distribution is Gaussian but one could output any parametric distribution by simply changing the loss function and the output nodes.

Our decoder consists of three main components: i) a target feed forward embedding layer ($tFF$), ii) a Cross-attention block ($CA$) and iii) an output layer ($O$). The $tFF$ layer maps the target queries to a set of continuous representations, the $CA$ block applies attention mechanism to align the target representations with encoder output, and the $O$ layer produces the mean and standard deviation of the target values.

The target embedding layer $tFF$ is a point wise feed forward layer that provides latent representation to the learned embedding of the target query. For a target query $w \in W$, we concatenate the time $t'$ and channel indicator $c'$ (one hot encoding), and get $\mathrm{w} \in \R^{C+1}$. $\mathrm{W} = \{\mathrm{w_1}, ..., \mathrm{w_r}\}$ is passed through $tFF$ to obtain their latent representation $Y^{(d)} = \{y_1^{(d)}, ..., y^{(d)}_r\}$.
The Cross-attention layer ($CA$) takes the $Y^{(d)}$ as query, and $Z^{(e)}$ as keys and values for a MAB, and outputs the learned embeddings $Z^{(d)} = \{z^{(d)}_1,..., z^{(d)}_r\}$. Finally, $Z^{(d)}$ is passed through the output layer $O$ which is a feed forward layer with two output heads. The output layer produces the distribution parameters $\mathrm{M}$ and $\Sigma$ corresponding to the elements present in $W$. We assume, for a predictor $w\in W$,  $\hat{Pr}(u'|X,w)$ defines the final probability distribution of $u'$ (target observation value) with $\hat{Pr}(u'|X,w) = \mathcal{N}(u'; \mu, \sigma)$.

The following equations describe the flow of data in the model. $E_{\Lambda}$ indicates embedding dimension of layer $\Lambda$ with $\Lambda \in \{iFF, SA, tFF, CA\}$.
\begin{align}
&	iFF: \mathrm{X} \mapsto Y^{(e)} \; \left( \in \R^{s \times E_{iFF}}\right)\\
&	SA: Y^{(e)} \mapsto Z^{(e)} \; \left(\in \R^{s \times E_{SA}}\right) \\
&	tFF: \mathrm{W} \mapsto Y^{(d)} \; \left(\in \R^{r \times E_{tFF}}\right) \\
&	CA: Y^{(d)}, Z^{(e)} \mapsto Z^{(d)} \; \left(\in \R^{r \times E_{CA}}\right)\\
&	O:  Z^{(d)} \mapsto \mathrm{M}, \Sigma \; \left(\in \R^r, {\R_+}^r \right) \\
& u' \sim \hat{Pr}(u'|X,w) = \mathcal{N}(u'; \mu, \sigma), \\
& \textnormal{where}  \quad w = (t',c') \in W, \quad \mu \in \mathrm{M}, \quad \sigma \in \Sigma \nonumber
\end{align}

In certain transformer models~\cite{VS17,ZZ21} for time series forecasting, the decoder employs a self-attention layer on target query embeddings. This introduces a transductive bias among independent variables (covariates). In our study, we lack covariates apart from observation time and channel. As a result, self-attention among target queries is unnecessary. Additionally, applying positional embeddings~\cite{VS17} to both encoder and decoder inputs is not needed. Our triplet already includes time and channel data, which naturally serves as positional embeddings for observations.


In the proposed Tripletformer, we use IMAB in encoder and MAB in the decoder. Although IMAB can be used for both Self-attention (in the encoder) and Cross-attention (in the decoder), we observed that using IMAB for Cross-attention is not providing consistent advantage over MAB in terms of prediction accuracy (see Section~\ref{sec:abl_tf_tf-dec-imab}).

\subsection{Supervised learning}

Our Tripletformer works in a heteroscedastic manner outputting probability distribution for each query. We assume that the data is following a Gaussian distribution, and use NLL as the main loss for training the model. In addition to NLL, we also use mean square error as the augmented loss in order to avoid the model sticking in local optima when the mean is almost flat after a few iterations. 
We optimize the following loss function $\mathcal{L}$:
\begin{align}
& \mathcal{L} = \sum_{n=1}^{N}-\E[\log \hat{Pr}({U'}^n|X^n, W^n)] + \lambda \E ||{U'}^n - \mathrm{M}^n||^2_2 \nonumber \\
& \textnormal{with} \quad \hat{Pr}({U'}^n|X^n,W^n) = \prod_{j=1}^{r^n} \hat{Pr}({u'}^n_j|X^n, w^n_j)
\end{align}

\section{Experiments}
\label{exp}

%

\begin{table}[ht]
	\centering
	\scriptsize
	\caption{Descrtiption of datasets used in our experiments. Sparsity indicates the $\%$ of missing observations when the time series is aligned.}
	\label{tab:data_desc}
	\begin{tabular}{lccccc}
		\toprule
		Dataset & \#Sample	&	\#Channel	& Avg. sparsity & Max.\#obs.	& Min.\#obs. \\
		\hline
		Physio'12 & 8,000 & 41 & 86 & 1154 & 29 \\
		MIMIC-III	& 21,000	& 17	& 67	& 1143 & 7 \\
		Physio'19 & 40,100 & 38 & 81 & 3080 & 13 \\
		PenDigits & 11,000 & 11 & 80 & 8& 8 \\
		Phon.Spec. & 6700 & 5 & 91 & 217 & 217 \\
		\hline
	\end{tabular}
\end{table}

\subsection{Datasets}
\label{sec:datasets}

We use the following real world IMTS datasets:
\textit{Physionet2012~\cite{SM12,GA00}}
Data from 8,000 ICU patients, covering up to 41 measurements taken within the first 48 hours after admission.
\textit{MIMIC-III~\cite{JP16}}
ICU records from around 21,000 stays, featuring irregularly sampled measurements and 17 observed variables. Dataset split using procedures from~\cite{HM20,HK19}.
\textit{Physionet2019~\cite{RJ19}}
For sepsis early detection, data from about 40,000 ICU samples across three U.S. hospitals, with 38 observed variables.

We also created two synthetic datasets from real world MTS namely PenDigits and PhonemeSpectra~\cite{RF21} (see Section~\ref{sec:exp_synth}) in order to verify the performance of the TripletFormer in extremely sparse scenario. We provide basic statistics of the datasets used for experiments in Table~\ref{tab:data_desc}.

\subsection{Competing Models}
\label{baselines}

Our Tripletformer competes with the following models. 
First, we compare with deterministic models that were made probabilistic by adding a homoscedastic variance which is a hyperparameter searched on the validation dataset.

\textit{Mean Imputation} always predicts the mean value of the channel in the training dataset. \textit{Forward Imputation} predicts the value of the previous observation in the channel. If there is no value observed in that particular channel, which is usual for IMTS, we impute with zero. \textit{L-ODE-RNN}~\cite{CR18} is a Neural Ordinary Different Equation (ODE) model where the decoder is an ODE-RNN while the encoder is made of an RNN. \textit{L-ODE-ODE}~\cite{RC19} is also an ODE model where the encoder consists of an ODE-RNN instead of an RNN. \textit{mTAN}~\cite{SM21}, Multi-Time attention Network that provides state-of-the-art results for deterministic interpolation, uses time attention networks for the series encoding and decoding. 

We also compare with the following probabilistic models:
\textit{GPR}~\cite{WR06} is the Gaussian Process regression modality where one model is trained per channel,
\textit{M-GPR~\cite{DP14}} is a multi-task Gaussian Process model where all the channels are trained simultaneously using a single architecture employing convolution of kernels,
\textit{HETVAE~\cite{SM22}}, Heteroscedastic Temporal Variational Autoencoder for Irregular Time Series (HETVAE) is the state-of-the-art probabilistic interpolation model for IMTS.
\textit{CSDI~\cite{TS21}}, a continuous score based model for probabilistic imputation of IMTS was evaluated for Continuous Ranked Probability Score (CRPS). Hence, we did a separate experiment to compare TripletFormer with CSDI for CRPS.

\begin{table}[t]
	\centering
	\scriptsize
	\caption{Results on real IMTS datasets, \textbf{observations missing at random}. Evaluation measure is NLL, lower the best. }
	\label{tab:random}
	\begin{tabular}{c|llll}
		\toprule
		& Obeserved $\%$ & $10\%$  & $50\%$ & $90\%$ \\
		\hline
		\hline
		&	Mean & 1.406{\tiny $\quad \quad \quad \, $} & 1.406{\tiny  $\quad \quad \quad \, $} & 1.402{\tiny  $\quad \quad \quad \, $}\\
		&	Forward & 1.371{\tiny  $\quad \quad \quad \, $} & 1.217{\tiny  $\quad \quad \quad \, $} & 1.164{\tiny  $\quad \quad \quad \, $}\\
 	&	GPR & 0.956{\tiny $\pm$0.001} & 0.767{\tiny $\pm$0.006} & 0.798{\tiny $\pm$0.004} \\
 	& M-GPR	& 1.249{\tiny $\pm$0.005}	& 1.191{\tiny $\pm$0.020} & 1.197{\tiny $\pm$0.055} \\
Physionet 	& L-ODE-RNN & 1.268{\tiny $\pm$0.012} & 1.233{\tiny $\pm$0.027} & 1.280{\tiny $\pm$0.029}
\\
2012 	& L-ODE-ODE & 1.211{\tiny $\pm$0.001}& 1.168{\tiny $\pm$0.001}& 1.170{\tiny $\pm$0.003} \\
		&	mTAN & 1.110{\tiny $\pm$0.000} & 0.934{\tiny $\pm$0.001} & 0.923{\tiny $\pm$0.002}\\
		&	HETVAE	& 0.849{\tiny $\pm$0.008}	& 0.578{\tiny $\pm$0.005}	&	0.551{\tiny $\pm$0.011}\\
		&	Tripletformer & \textbf{0.780$\pm${\tiny 0.013}} & \textbf{0.455{\tiny $\pm$0.011}} & \textbf{0.373{\tiny $\pm$0.040}} \\
		\cline{2-5}
		& \% Improvement & 8.1\%	&	21.2\%	&	32.3\%	\\
		\hline
		\hline
		\multirow{7}{*}{MIMIC-III}
		&	Mean & 1.508{\tiny  $\quad \quad \quad \, $} & 1.508{\tiny  $\quad \quad \quad \, $} & 1.507{\tiny $\quad \quad \quad \, $}\\
		&	Forward & 1.750{\tiny  $\quad \quad \quad \, $}& 1.423{\tiny  $\quad \quad \quad \, $} &1.328 {\tiny  $\quad \quad \quad \, $} \\
		&	GPR & 1.201{\tiny $\pm$0.003} & 0.979{\tiny $\pm$0.006} &  0.919{\tiny $\pm$0.001} \\
		& M-GPR & 1.695{\tiny $\pm$0.041}	& 1.302{\tiny $\pm$0.068} & 1.297{\tiny $\pm$0.121} \\
		&	mTAN & 1.209{\tiny $\pm$0.000} & 1.066{\tiny $\pm$0.001} & 1.065{\tiny $\pm$0.001}\\
		&	HETVAE	& 1.077{\tiny $\pm$0.003}	& 0.828{\tiny $\pm$0.001}	&	0.774{\tiny $\pm$0.008} \\
		&	Tripletformer & \textbf{1.056{\tiny $\pm$0.006}} & \textbf{0.789{\tiny $\pm$0.006}} & \textbf{0.710{\tiny $\pm$0.010}} \\\cline{2-5}
		& \% Improvement & 1.9\%	&	4.7\%	&	8.3\%	\\
		\hline
		\hline
		&	Mean & 1.421{\tiny  $\quad \quad \quad \, $} & 1.420{\tiny  $\quad \quad \quad \, $} &1.425{\tiny  $\quad \quad \quad \, $}\\
		&	Forward & 1.361{\tiny  $\quad \quad \quad \, $} & 1.205{\tiny  $\quad \quad \quad \, $} & 1.433{\tiny  $\quad \quad \quad \, $}  \\
	Physionet	&	GPR & 1.136{\tiny $\pm$0.000} &  0.907{\tiny $\pm$0.012} & 0.851{\tiny $\pm$0.007} \\
	2019	& 	mTAN &  1.152{\tiny $\pm$0.001} & 0.988{\tiny $\pm$0.002} & 0.982{\tiny $\pm$0.006}\\
		&	HETVAE	& 1.091{\tiny $\pm$0.001}& 0.855{\tiny $\pm$0.002} & 0.835{\tiny $\pm$0.005}\\
		&	Tripletformer & \textbf{1.079{\tiny $\pm$0.001}} & \textbf{0.806{\tiny $\pm$0.004}} &  \textbf{0.752{\tiny $\pm$0.007}}\\
		\cline{2-5}
		& \% Improvement & 1.1\%	&	5.7\%	&	9.9\%	\\
		\bottomrule
		
	\end{tabular}
\end{table}

\begin{table}[t]
	\centering
	\scriptsize
	\caption{Results of \textbf{bursts of observations missing} on real IMTS datasets. Evaluation measure is NLL, lower the best.}
	\label{tab:burst}
	\begin{tabular}{c|llll}
		
		\toprule
		&	Obeserved $\%$ & $10\%$  & $50\%$ & $90\%$ \\
		\hline
		\hline
		&	Mean & 1.412{\tiny  $\quad \quad \quad \, $} & 1.382{\tiny  $\quad \quad \quad \, $} & 1.407{\tiny  $\quad \quad \quad \, $}\\
		&	Forward & 1.534{\tiny  $\quad \quad \quad \, $}& 1.442{\tiny  $\quad \quad \quad \, $} & 1.232{\tiny  $\quad \quad \quad \, $}\\
		&	GPR & 1.127{\tiny $\pm$0.018} & 0.983{\tiny $\pm$0.002} & 0.691{\tiny $\pm$0.001} \\
		& M-GPR & 1.351{\tiny $\pm$0.023} & 1.255{\tiny $\pm$0.037} & 1.221{\tiny $\pm$0.030} \\
Physionet	& L-ODE-RNN & 1.270{\tiny $\pm$0.002}& 1.382{\tiny $\pm$0.006}& 1.289{\tiny $\pm$0.011}\\
2012	& L-ODE-ODE & 1.263{\tiny $\pm$0.000}& 1.242{\tiny $\pm$0.002}& 1.243{\tiny $\pm$0.001}\\
		&	mTAN & 1.200{\tiny $\pm$0.000} & 1.141{\tiny $\pm$0.001} & 0.960{\tiny $\pm$0.001}\\
		&	HETVAE	& 1.028{\tiny $\pm$0.011} & 0.882{\tiny $\pm$0.002} & {0.614{\tiny $\pm$0.039}}\\
		&	Tripletformer & \textbf{0.925{\tiny $\pm$0.005}}& \textbf{0.777{\tiny $\pm$0.013}} & \textbf{0.578{\tiny $\pm$0.034}}\\
		\cline{2-5}
		& \% Improvement & 11.1\%	&	11.9\%	&	5.8\%	\\
		\hline
		\hline
		\multirow{7}{*}{MIMIC-III}
		&	Mean & 1.509{\tiny  $\quad \quad \quad \, $}&1.507{\tiny  $\quad \quad \quad \, $} & 1.515{\tiny  $\quad \quad \quad \, $}\\
		&	Forward & 1.870{\tiny  $\quad \quad \quad \, $} & 1.608{\tiny  $\quad \quad \quad \, $} & 1.412{\tiny  $\quad \quad \quad \, $} \\
		&	GPR & 1.328{\tiny $\pm$0.000} & 1.231{\tiny $\pm$0.004} & 1.000{\tiny $\pm$0.039} \\
		& M-GPR & 1.637{\tiny $\pm$0.08} & 1.42{\tiny $\pm$0.131} & 1.535{\tiny $\pm$0.138} \\
		&	mTAN & 1.319{\tiny $\pm$0.001} & 1.247{\tiny $\pm$0.000} & 1.094{\tiny $\pm$0.002} \\
		&	HETVAE	& 1.210{\tiny $\pm$0.000}& 1.126{\tiny $\pm$0.003} & 0.939{\tiny $\pm$0.009}\\
		&	Tripletformer & \textbf{1.193{\tiny $\pm$0.003}}& \textbf{1.087{\tiny $\pm$0.006}} &  \textbf{0.885{\tiny $\pm$0.032}}\\
		\cline{2-5}
		& \% Improvement & 1.4\%	&	3.6\%	&	5.8\%	\\
		\hline
		\hline
		&	Mean & 1.425{\tiny  $\quad \quad \quad \, $} & 1.415{\tiny  $\quad \quad \quad \, $} &1.419{\tiny  $\quad \quad \quad \, $}\\
		&	Forward & 1.466{\tiny  $\quad \quad \quad \, $} & 1.368{\tiny  $\quad \quad \quad \, $} &  1.209{\tiny  $\quad \quad \quad \, $} \\
	Physionet	&	GPR & 1.257{\tiny $\pm$0.001} & 1.049{\tiny $\pm$0.002}  & 0.975{\tiny $\pm$0.005} \\
	2019	& 	mTAN & 1.269{\tiny $\pm$0.001} & 1.113{\tiny $\pm$0.001} & 1.042{\tiny $\pm$0.002}\\
		&	HETVAE	& 1.241{\tiny $\pm$0.001}& 1.012{\tiny $\pm$0.004} & 0.949{\tiny $\pm$0.002}\\
		&	Tripletformer & \textbf{1.222{\tiny $\pm$0.001}}& \textbf{0.977{\tiny $\pm$0.005}} & \textbf{0.865{\tiny $\pm$0.009}} \\
		\cline{2-5}
		& \% Improvement & 1.5\%	&	3.5\%	&	8.9\%	\\
		\bottomrule
		
	\end{tabular}
\end{table}

\subsection{Experimental Setup}
\label{sec:exp_setup}

We randomly split each dataset into train and test with $80\%$ and $20\%$ samples. Further $20\%$ of the train set is used as validation set. We search the following hyperparameters: $L \in \{1,2,3,4\}$, hidden units in
MLP layers (feed forward layers in both encoder and decoder) from $\{64,128,256\}$, attention layers dimension from $\{64,128,256\}$, number of induced points in IMAB from $\{16,32,64,128\}$, and $\lambda$ augmented loss weight from $\{0,1,5,10\}$. Similar to~\cite{HM20}, we trained all the competing models for $5$ different random hyperparameter setups, and chose the one that has the least NLL on the validation dataset. We independently run the experiments for $5$ times using the best hyperparameters and different random seeds. 
We code all the models using PyTorch, and trained on NVIDIA GeForce RTX 3090 GPUs.

\subsubsection*{Data preprocessing} 
\label{sec:apd_data_prepro}

We found that some observations in the datasets contain errors. For example, in the Physionet dataset, we noticed that some of the Ph Value readings were around $700$ which is outside the normal range of $0$ to $14$. To address this issue, we implemented a standard practice for handling real-world IMTS datasets by identifying and removing observations that fall outside the $99.9^{th}$ percentile of the observation range in the training data. To further preprocess the data, we also rescaled the time values between $0$ and $1$ for all datasets and standardized each channel to the standard normal distribution.



\subsection{Task setting}
\label{exp:sam_tech}

In the experiments, we considered two types of setups for the interpolation task.

\subsubsection*{Random missing} In this setup, we assume that the sensors are missed at random time points. Hence we condition on the random time points and predict the observation values of all the available channels for the remaining time points.

\subsubsection*{Burst missing} Here, we assume that the sensors are not observed for a series of time points. We randomly choose a start time point, and use $p$ many time points after that for prediction by conditioning upon the remaining.

\subsection{Results on real IMTS datasets}
\label{sec:exp_results}

Our main experimental results using two different sampling techniques on real IMTS datasets are presented below.

\subsubsection{Results on random missing}

In Table~\ref{tab:random}, we present the results for the random missing setup. Three different conditioning ranges (observed $\%$) which are $10\%, 50\%$ and $90\%$ of all the observations in the series were used. We could not run ODE models on MIMIC-III and Physionet2019 datasets due to high computational requirement. However, for the datasets where they have been evaluated, ODE models performed significantly worse than Tripletformer. Hence, it is safe to assume that ODE models perform worse in the remaining datasets as well. Similarly, we could not compute M-GPR on Physionet2019 due to limited computational resources as it requires computing the inverse of co-variance matrix.

As also observed in~\cite{SM22}, we notice that M-GPR performs worse than GPR.  We see that our Tripletformer outperforms all the competing models with a huge margin. Especially, compared to the next best model, HETVAE, significant lifts are observed in the Physionet2012 dataset. 

We note that the published results~\cite{SM22} with $50\%$ time points in conditioning range using random missing for both the Physionet2012 and MIMIC-III differ from ours. For the Physionet2012, this can be attributed to the following: 1) in~\cite{SM22}, the dataset was normalized using the parameters computed on the entire dataset instead of training dataset, 2) inclusion of erroneous observations in~\cite{SM22} (see Section~\ref{sec:exp_setup}). Also, we were not able to reproduce the results with the preprocessing details provided in~\cite{SM22}. Whereas, for MIMIC-III dataset, we could not extract the splits used in~\cite{SM22}, hence used another standard version~\cite{HM20, HK19}.

We would like to note that the proposed Tripletformer is scalable to the large time series datasets. IMAB consists of near-linear computational complexity as explained in Section~\ref{sec:imab} making the model scalable. As an example, Tripletformer can seamlessly process Physionet2019 dataset which has up to 3080 observations.

\begin{figure}[t]
	\centering
	\subfigure[Physionet2012, random]{
		\includegraphics[clip, width=0.45\columnwidth]{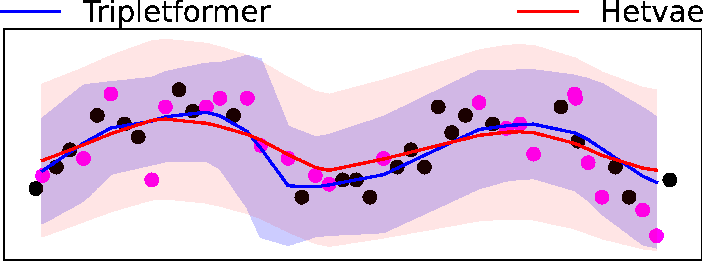}
	}
	\hfill
	\subfigure[Physionet2012, burst]{
		\includegraphics[clip, width=0.45\columnwidth]{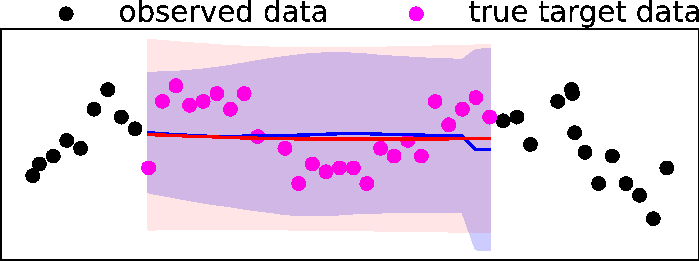}
	}
	\subfigure[MIMIC-III, random]{
		\includegraphics[clip, width=0.45\columnwidth]{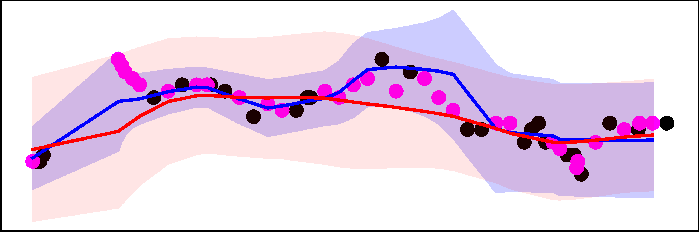}
	}
	\hfill
	\subfigure[MIMIC-III, burst]{
		\includegraphics[clip, width=0.45\columnwidth]{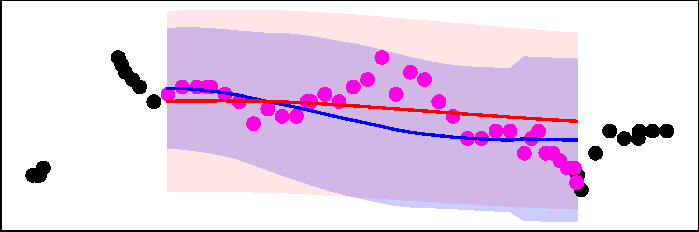}
	}
	\subfigure[Physionet2019, random]{
		\includegraphics[clip, width=0.45\columnwidth]{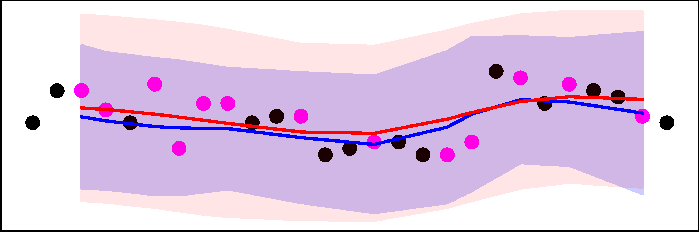}
	}
	\hfill
	\subfigure[Physionet2019, burst]{
		\includegraphics[clip, width=0.45\columnwidth]{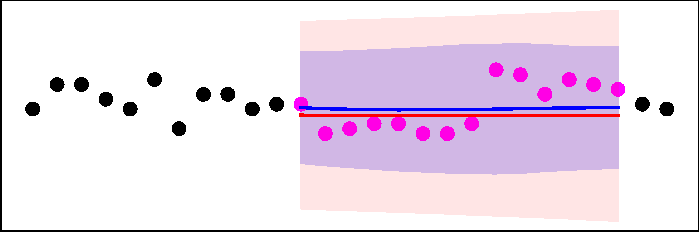}
	}
	\caption{Comparison of qualitative performance between Tripletformer and HETVAE. Plots are the predictions ($95^{th}$ quantile) of both the models for heart rate in all the three datasets conditioned on the $50\%$ of the time points.}
	\label{fig:qual_comp}
\end{figure}

\subsubsection{Results on burst missing}
We present the results for burst sampling in Table~\ref{tab:burst}. Interpolating the burst missing values is a difficult task compared to that of random missing. Because, in burst missing, conditioning time points are far from the query time point whereas in random sampling they are close by.
The results on burst sampling show a similar pattern to random sampling. Our Tripletformer outperforms all the models with a significant margin; HETVAE and GPR are the second and third best models. 
We see significant lifts in this setup as well compared to the state-of-the-art model, HETVAE. Again, in Physionet2012 dataset with $50\%$ of the observations in the conditioning range, our Tripletformer has around $12\%$ improvement in NLL.

We perform a qualitative comparison of the predictions made by the Tripletformer and HETVAE models in Figure~\ref{fig:qual_comp}. We observe that the Tripletformer provides not only accurate predictions but also with higher certainty which is desirable in probabilistic interpolation.

\begin{table}[t]
	\centering
	\scriptsize
	\caption{Results on synthetic IMTS, \textbf{observations missing at random}. Evaluation measure is NLL, lower the best. }
	\label{tab:synth:random}
	\begin{tabular}{c|l|ccc}
		\toprule
		
		&	Obeserved $\%$ & $10\%$  & $50\%$ & $90\%$ \\
		\hline
		
		\hline
		&	Mean & 1.424{\tiny  $\quad \quad \quad \, $} & 1.420{\tiny  $\quad \quad \quad \, $} & 1.424{\tiny  $\quad \quad \quad \, $}\\
		
		&	Forward & 1.550{\tiny  $\quad \quad \quad \, $} & 1.626{\tiny  $\quad \quad \quad \, $} & 1.596{\tiny  $\quad \quad \quad \, $}\\
		
		&	GPR & 1.411{\tiny $\pm$0.001} &1.350{\tiny $\pm$0.030} & 1.279{\tiny $\pm$0.055}\\
		& M-GPR & 1.400{\tiny $\pm$0.001} & 1.326{\tiny $\pm$0.059} & 1.558{\tiny $\pm$0.474} \\
		
		Pen &	L-ODE-RNN & 1.285{\tiny $\pm$0.005}& 1.183{\tiny $\pm$0.012}& 1.102{\tiny $\pm$0.014} \\
		
		Digits &	L-ODE-ODE & 1.307{\tiny $\pm$0.001}& 1.182{\tiny $\pm$0.002} & 1.108{\tiny $\pm$0.004} \\
		
		&	mTAN & 1.307{\tiny $\pm$0.001} & 1.050{\tiny $\pm$0.013} & 0.882{\tiny $\pm$0.020} \\
		
		&	HETVAE & 1.267{\tiny $\pm$0.003} & 1.207{\tiny $\pm$0.025} & 1.326{\tiny $\pm$0.015}\\
		
		&	Tripletformer & \textbf{1.115{\tiny $\pm$0.003}}& \textbf{0.693{\tiny $\pm$0.019}} & \textbf{0.463{\tiny $\pm$0.027}}\\
		\cline{2-5}
		& \% Improvement & 11.9\%	&	42.6\%	&	65.1\%	\\
		\hline
		
		\hline
		&	Mean & 1.437{\tiny  $\quad \quad \quad \, $} & 1.437{\tiny  $\quad \quad \quad \, $} & 1.435{\tiny  $\quad \quad \quad \, $}\\
		&	Forward & 1.553{\tiny  $\quad \quad \quad \, $} & 1.559{\tiny  $\quad \quad \quad \, $} & 1.528{\tiny  $\quad \quad \quad \, $}\\
		&	GPR & 1.378{\tiny $\pm$0.002} & 1.271{\tiny $\pm$0.005} & 1.176{\tiny $\pm$0.009}\\
		& M-GPR & 1.301{\tiny $\pm$0.007} & 1.116{\tiny $\pm$0.077} & 1.961{\tiny $\pm$1.385} \\
		Phoneme & L-ODE-RNN & 1.327{\tiny $\pm$0.006}& 1.289{\tiny $\pm$0.010}& 1.263{\tiny $\pm$0.025}\\
		Spectra & L-ODE-ODE & 1.304{\tiny $\pm$0.002}& 1.273{\tiny $\pm$0.006}& 1.264{\tiny $\pm$0.008} \\
		&	mTAN & 1.225{\tiny $\pm$0.003} & 0.963{\tiny $\pm$0.009} & 1.097{\tiny $\pm$0.010}\\
		&	HETVAE	& 1.033{\tiny $\pm$0.031} & 0.699{\tiny $\pm$0.004} & 0.804{\tiny $\pm$0.002}\\
		&	Tripletformer &\textbf{0.923{\tiny $\pm$0.008}} & \textbf{0.413{\tiny $\pm$0.005}} & \textbf{0.115{\tiny $\pm$0.024}}\\
		\cline{2-5}
		& \% Improvement & 10.7\%	&	40.9\%	&	85.7\%	\\
		\bottomrule
	\end{tabular}
\end{table}

\begin{table}[t]
	\centering
	\scriptsize
	\caption{Results of synthetic IMTS for \textbf{observations missing at burst}. Evaluation measure is NLL, lower the best. }
	\label{tab:synth:burst}
	\begin{tabular}{c|l|ccc}
		\toprule
		
		&	Obeserved $\%$ & $10\%$  & $50\%$ & $90\%$ \\
		\hline
		
		\hline
		
		&	Mean  & 1.427{\tiny  $\quad \quad \quad \, $} & 1.396{\tiny  $\quad \quad \quad \, $} & 1.424{\tiny  $\quad \quad \quad \, $}\\
		&	Forward & 1.556{\tiny  $\quad \quad \quad \, $} & 1.683{\tiny  $\quad \quad \quad \, $} & 1.596{\tiny  $\quad \quad \quad \, $}\\
		&	GPR & 1.386{\tiny $\pm$0.006} & 1.413{\tiny $\pm$0.000} & 1.279{\tiny $\pm$0.055} \\
		& M-GPR & 1.381{\tiny $\pm$0.010}	& 1.345{\tiny $\pm$0.012} & 1.344{\tiny $\pm$0.218} \\
		Pen &	L-ODE-RNN & 1.271{\tiny $\pm$0.002}& 1.184{\tiny $\pm$0.005}& 1.144{\tiny $\pm$0.007} \\
		Digits &	L-ODE-ODE & 1.269{\tiny $\pm$0.002}& 1.207{\tiny $\pm$0.020} & 1.171{\tiny $\pm$0.075}		\\
		&	mTAN & 1.277{\tiny $\pm$0.002} & 1.061{\tiny $\pm$0.004} & 0.882{\tiny $\pm$0.023}\\
		&	HETVAE & 1.270{\tiny $\pm$0.001} & 1.188{\tiny $\pm$0.006} & 1.326{\tiny $\pm$0.033}\\
		&	Tripletformer & \textbf{1.115{\tiny $\pm$0.000}} & \textbf{0.787{\tiny $\pm$0.015}}		 & \textbf{0.463\tiny{$\pm$ 0.027}}\\
		\cline{2-5}
		& \% Improvement & 12.2\%	&	33.8\%	&	65.1\%	\\
		\hline
		
		\hline
		&	Mean & 1.441{\tiny  $\quad \quad \quad \, $} & 1.436{\tiny  $\quad \quad \quad \, $} & 1.456{\tiny  $\quad \quad \quad \, $}\\
		&	Forward & 1.583{\tiny  $\quad \quad \quad \, $} & 1.643{\tiny  $\quad \quad \quad \, $} & 1.610{\tiny  $\quad \quad \quad \, $}\\
		&	GPR & 1.387{\tiny $\pm$0.000} & 1.331{\tiny $\pm$0.000} & 1.309{\tiny $\pm$0.001}\\
		& M-GPR & 1.469{\tiny $\pm$0.101} & 1.345{\tiny $\pm$0.035} & 1.305{\tiny $\pm$0.011} \\
		Phoneme &	L-ODE-RNN & 1.389{\tiny $\pm$0.010}& 1.356{\tiny $\pm$0.013}& 1.309{\tiny $\pm$0.002} \\
		Spectra &	L-ODE-ODE & 1.367{\tiny $\pm$0.001}& 1.343{\tiny $\pm$0.002}& 1.290{\tiny $\pm$0.001}\\
		&	mTAN & 1.350{\tiny $\pm$0.001} & 1.285{\tiny $\pm$0.001} & 1.314{\tiny $\pm$0.008}\\
		&	HETVAE	& 1.237{\tiny $\pm$0.003} & 1.057{\tiny $\pm$0.002} & 1.006{\tiny $\pm$0.002}\\
		&	Tripletformer &\textbf{1.180{\tiny $\pm$0.008}} & \textbf{1.025{\tiny $\pm$0.005}} & \textbf{0.975{\tiny $\pm$0.009}}\\
		\cline{2-5}
		& \% Improvement & 4.6\%	&	3.2\%	&	3.1\%	\\
		\bottomrule
		
	\end{tabular}
\end{table}

\subsection{Experiments with synthetic IMTS datasets}
\label{sec:exp_synth}

In this experiment, we see the performance of the competing models in the extremely sparse setup. We set MTS to be observed asynchronously meaning each senor is observed independent of others. Hence, we assume that at every point of time only one sensor is observed. From MTS, we randomly chose one variable at a single point of time, and remove all the remaining variables making the number of observations in synthetic IMTS is same as the length of the source MTS.

For this, we choose PenDegits and PhonemeSpectra which are the second and third largest datasets used in~\cite{RF21}. While the largest dataset, FaceDetection, lacked variability in interpolation results across models and sampling techniques, PenDigits and PhonemeSpectra were chosen as suitable alternatives. Experimental results for random and burst sampling are presented in  Tables~\ref{tab:synth:random} and~\ref{tab:synth:burst} respectively. For PenDigits, both sampling techniques yielded the same results due to the short time series length when $90\%$ of the series is observed. For both the sampling types, and all the conditioning ranges, Tripletformer provides superior interpolations.

The reason for the poor performance of the HETVAE on synthetic datasets is the encoding of each channel separately. We note that the medical datasets do not have significant cross channel interactions compared to the synthetic IMTS datasets making HETVAE shine comparatively better in the former. However, {\em for both synthetic and real IMTS datasets, Tripletformer outperforms HETVAE significantly}.

\subsection{Comparison with CSDI~\cite{TS21}}
\label{sec:csdi_comp}

CSDI is a score based diffusion model that produce probabilistic outputs. Here, we compare the Tripletformer with CSDI in terms of Continuous Ranked Probability Score (CRPS) as shown in~\cite{TS21}. Because of huge computational requirement of diffusion models, conducting experiments with CSDI on all the datasets is beyond our computational resources, hence, we compare both Tripletformer and HETVAE on the published results from~\cite{TS21} in Table~\ref{tab:csdi_phy}. For fair comparison, we use the same data splits of Physionet2012 data that was given in~\cite{TS21}. We note that CSDI produces outputs with homoscedastic variance similar to that of mTAN and L-ODE-ODE. It an be observed that the Tripletformer outperforms all the baseline models by significant margin for the CRPS score as well. On an average, {\em Tripletformer improves the interpolation accuracy by $63\%$ and $86\%$ respectively compared to HETVAE and CSDI models respectively}.

\begin{table}[t]
	\scriptsize
	\centering
	\caption{Comparison of Tripletformer and HETVAE with published results from~\cite{TS21} on Physionet2012 dataset and random missing. Evaluation measure is CRPS, lower the best. $\dagger$ indicates published results from ~\cite{TS21}.}
	\label{tab:csdi_phy}
	\begin{tabular}{lccc}
		\hline 
		Observed$\%$ & $10 \%$ & $50 \%$ & $90 \%$ \\
		\hline
		L-ODE-ODE & 0.761{\tiny $\pm$0.010}${^\dagger}$ & 0.676{\tiny $\pm$0.003}${^\dagger}$ & 0.700{\tiny $\pm$0.002}${^\dagger}$ \\
		mTAN & 0.689{\tiny $\pm$0.015}${^\dagger}$ & 0.567{\tiny $\pm$0.003}${^\dagger}$ & 0.526{\tiny{$\pm$0.004}}${^\dagger}$ \\
		HETVAE & 0.188{\tiny $\pm$0.007} & 0.116{\tiny $\pm$0.016}& 0.256{\tiny $\pm$0.003} \\
		CSDI & 0.556{\tiny $\pm$0.003}${^\dagger}$ & 0.418{\tiny $\pm$0.001}${^\dagger}$ & 0.380{\tiny $\pm$0.002}${^\dagger}$ \\
		Tripletformer & \textbf{0.062{\tiny $\pm$0.006}} & \textbf{0.062{\tiny $\pm$0.009}} & \textbf{0.061{\tiny $\pm$0.010}} \\
		\hline
	\end{tabular}
\end{table}

\subsection{Experiment on Deterministic Interpolation}
\label{sec:det_int}

\begin{table}[ht]
	\scriptsize
	\centering
	\caption{Comparison of competing models for deterministic interpolation on Physionet2012 dataset and random missing. Evaluation metric Mean Squared Error, lower the best. $\dagger$ indicates published results from ~\cite{SM21}.}
	\label{tab:det_phy}
	\begin{tabular}{lccc}
		\hline 
		Model & \multicolumn{3}{c}{ Mean Squared Error $\left(\times 10^{-3}\right)$} \\
		\hline
		L-ODE-RNN & 8.132{\tiny $\pm$0.020}$^{\dagger}$ & 8.171{\tiny $\pm$0.030}$^{\dagger}$ & 8.402{\tiny $\pm$0.022}$^{\dagger}$ \\
		L-ODE-ODE & 6.721{\tiny $\pm$0.109}$^{\dagger}$ & 6.798{\tiny $\pm$0.143}$^{\dagger}$ & 7.142{\tiny $\pm$0.066}$^{\dagger}$ \\
		mTAN & 4.139{\tiny $\pm$0.029}$^{\dagger}$ & 4.157{\tiny $\pm$0.053}$^{\dagger}$ & 4.798{\tiny{$\pm$0.036}}$^{\dagger}$ \\
		HETVAE & 4.200{\tiny $\pm$0.500} & 4.945{\tiny $\pm$0.000}& 4.600{\tiny $\pm$0.000} \\
		Tripletformer & \textbf{3.500{\tiny $\pm$0.100}} & \textbf{3.600{\tiny $\pm$0.100}} & \textbf{3.900{\tiny $\pm$0.200}} \\
		\hline Observed $\%$ & $50 \%$ & $70 \%$ & $90 \%$ \\
		\hline
	\end{tabular}
\end{table}

It is interesting to see the performance of Tripletformer for the deterministic interpolation. We trained both Tripletformer and HETVAE for predicting the mean value (optimized for Mean Squared Error loss). Tripletformer is compared with HETVAE~\cite{SM22} (probabilistic model), mTAN~\cite{SM21}, L-ODE-ODE~\cite{RC19} and L-ODE-RNN~\cite{CR18} models on Physionet2012 dataset for random missing setup in Table~\ref{tab:det_phy}. While we did an experiment for HETVAE, published results from~\cite{SM21} were reported for mTAN and ODE models because we use the same dataset splits provided by~\cite{SM21}. Notably {\em Tripletformer outperforms all the models with a significant margin for deterministic interpolation as well, and reduced the mean squared error of the next best model by $12\%$ on an average}.

\subsection{Ablation Study}
\label{sec:exp_abl}

\subsubsection{Tripletformer vs. TF-enc-MAB}

In table~\ref{tab:abl1}, we compare the performance of Tripletformer with its variant Tf-enc-MAB which  consists of MAB in the encoder. We use synthetic IMTS datasets for comparison because we could not run Tf-enc-MAB on all the splits of real IMTS datasets.
We see that choosing IMAB instead of MAB in the encoder is optimal. As mentioned in Section~\ref{sec:imab}, Tripletformer that has IMAB in the encoder provides similar or better performance compared to Tf-enc-MAB demonstrating the advantage of restricted attention in IMAB. We see significant lifts when conditioning range increases from $10\%$ to $90\%$.



\begin{table}[ht]
	\centering
	\scriptsize
	\caption{Comparing Tripletformer and Tf-enc-MAB. Tf-enc-MAB is a variant of Tripletformer and consists of the MAB in the encoder instead of the IMAB.
		Results of synthetic IMTS. Evaluation measure is NLL, lower the best. }
	\label{tab:abl1}
	\begin{tabular}{l|l|ccc}
		\toprule
		
	&	tp. obs. & $10\%$  & $50\%$ & $90\%$ \\
		\hline
		\multicolumn{5}{c}{random sampling} \\
		\hline
	Pen &	Tf-enc-MAB & 1.117{\tiny $\pm$0.001} & 0.713{\tiny $\pm$0.009} & 0.551{\tiny $\pm$0.026}\\
	Digits&	Tripletformer & \textbf{1.115{\tiny $\pm$0.003}}& \textbf{0.693{\tiny $\pm$0.019}} & \textbf{0.463{\tiny $\pm$0.027}}\\
		
		\hline
		
	Phoneme &	Tf-enc-MAB & 0.966{\tiny $\pm$0.010} & 0.626{\tiny $\pm$0.142} & 0.504{\tiny $\pm$0.156}\\
	Spectra &	Tripletformer & \textbf{0.923{\tiny $\pm$0.008}} & \textbf{0.413{\tiny $\pm$0.005}} & \textbf{0.115{\tiny $\pm$0.024}}\\
		\hline
		
				\multicolumn{5}{c}{burst sampling} \\
		\hline
		Pen &	Tf-enc-MAB & 1.111{\tiny $\pm$0.002} & 0.838{\tiny $\pm$0.020} & 0.551{\tiny $\pm$0.026}\\
		Digits&	Tripletformer & \textbf{1.115{\tiny $\pm$0.000}} & \textbf{0.787{\tiny $\pm$0.015}}		 & \textbf{0.463\tiny{$\pm$ 0.027}}\\
		
		\hline
		
		\hline
		Phoneme &	Tf-enc-MAB & 1.205{\tiny $\pm$0.009}& 1.035{\tiny $\pm$0.004}& 1.019{\tiny $\pm$0.026}\\
		Spectra &	Tripletformer & \textbf{1.180{\tiny $\pm$0.008}} & \textbf{1.025{\tiny $\pm$0.005}} & \textbf{0.975{\tiny $\pm$0.009}} \\
		\bottomrule
		
	\end{tabular}
\end{table}

\subsubsection{Tripletformer vs. Tf-dec-IMAB}
\label{sec:abl_tf_tf-dec-imab}

We compare the Tripletformer with Tf-dec-IMAB which is a variant of Tripletformer where MAB is replaced with IMAB in the decoder. Comparisons are made using all the datasets for both random and burst sampling and the results are presented in Table~\ref{tab:abl2}. We see that, among $17$ out of $30$ comparisons Tripletformer performs better. Among real IMTS datasets, Tf-dec-IMAB provides better performance in burst sampling for MIMIC-III with $50\%$ and $90\%$ time points and Physionet2019 in $90\%$ time points in conditioning range. Whereas Tripletformer has better performance in Physionet2012 dataset for both sampling techniques when conditioned up of $50\%$ and $90\%$ of the available time points. We see similar behavior for synthetic IMTS datasets as well. While Tf-dec-IMAB performs better for PenDigits, Tripletformer performs better in PhonemeSpectra. We see that though Tf-dec-IMAB predicts well in some scenarios, on an average Tripletformer provide around $5\%$ less error.

\begin{table}[ht]
	\centering
	\scriptsize
	\caption{Comparing Tripletformer and Tf-dec-MAB (variant of Tripletformer, consists the IMAB in the decoder instead of the MAB). Evaluation measure is NLL, lower the best. }
	\label{tab:abl2}
	\begin{tabular}{l|l|ccc}
		\toprule
		
		&	tp. obs. & $10\%$  & $50\%$ & $90\%$ \\
		\hline
		\multicolumn{5}{c}{random sampling} \\
		
		\hline
		\multirow{2}{*}{Physionet'12} &	Tf-dec-IMAB &  0.845{\tiny $\pm$0.015}& 0.565{\tiny $\pm$0.05} & 0.703{\tiny $\pm$0.092}\\ 
		&	Tripletformer & \textbf{0.780$\pm${\tiny 0.013}}& \textbf{0.455{\tiny $\pm$0.011}} & \textbf{0.373{\tiny $\pm$0.040}}\\
		\hline
		\multirow{2}{*}{MIMIC-III} &	Tf-dec-IMAB &  \textbf{1.053{\tiny $\pm$0.037}} & \textbf{0.783{\tiny $\pm$0.01}} & 0.735{\tiny $\pm$0.018} \\ 
		&	Tripletformer & 1.056{\tiny $\pm$0.006} & {0.789{\tiny $\pm$0.006}} & \textbf{0.710{\tiny $\pm$0.010}}\\
		\hline
		\multirow{2}{*}{Physionet'19} &	Tf-dec-IMAB &  \textbf{1.061{\tiny $\pm$0.003}} & \textbf{0.796{\tiny $\pm$0.002}} & 0.797{\tiny $\pm$0.017} \\ 
		&	Tripletformer & 1.079{\tiny $\pm$0.001} & 0.806{\tiny $\pm$0.004} & \textbf{0.752{\tiny $\pm$0.007}}\\
		
		\hline
		Pen &	Tf-dec-IMAB &  1.113{\tiny $\pm$0.001}& 0.717{\tiny $\pm$0.028} & \textbf{0.386{\tiny $\pm$0.054}}\\ 
		Digits&	Tripletformer & \textbf{1.115{\tiny $\pm$0.003}}& \textbf{0.693{\tiny $\pm$0.019}} & {0.463{\tiny $\pm$0.027}}\\
		
		\hline
		
		Phoneme &	Tf-dec-IMAB & 0.951{\tiny $\pm$0.009} & 0.570{\tiny $\pm$0.174} & 0.249{\tiny $\pm$0.042}\\
		Spectra &	Tripletformer & \textbf{0.923{\tiny $\pm$0.008}} & \textbf{0.413{\tiny $\pm$0.005}} & \textbf{0.115{\tiny $\pm$0.024}}\\
		\hline
		
		\multicolumn{5}{c}{burst sampling} \\
		
				\hline
		\multirow{2}{*}{Physionet'12} &	Tf-dec-IMAB &  0.946{\tiny $\pm$0.007} & 0.821{\tiny $\pm$0.008} & 0.794{\tiny $\pm$0.023} \\ 
		&	Tripletformer & \textbf{0.925{\tiny $\pm$0.005}} & \textbf{0.777{\tiny $\pm$0.013}} & \textbf{0.578{\tiny $\pm$0.034}} \\
		\hline
		\multirow{2}{*}{MIMIC-III} &	Tf-dec-IMAB &  1.203{\tiny $\pm$0.003} & \textbf{1.023{\tiny $\pm$0.006}} & \textbf{0.835{\tiny $\pm$0.015}} \\ 
		&	Tripletformer & \textbf{1.193{\tiny $\pm$0.003}} & {1.087{\tiny $\pm$0.006}} & {0.885{\tiny $\pm$0.032}}\\
		\hline
		\multirow{2}{*}{Physionet'19} &	Tf-dec-IMAB &  \textbf{1.210{\tiny $\pm$0.001}} & 0.980{\tiny $\pm$0.003} & 0.797{\tiny $\pm$0.017} \\ 
		&	Tripletformer & {1.222{\tiny $\pm$0.001}} & \textbf{0.977{\tiny $\pm$0.005}} & 0.865{\tiny $\pm$0.009} \\
		
		\hline
		Pen &	Tf-dec-IMAB &  1.127{\tiny $\pm$0.001} & \textbf{0.735{\tiny $\pm$0.015}} & \textbf{0.386{\tiny $\pm$0.054}}\\ 
		Digits&	Tripletformer & \textbf{1.115{\tiny $\pm$0.000}} & {0.787{\tiny $\pm$0.015}} & {0.463\tiny{$\pm$ 0.027}}\\
		
		\hline
		
		\hline
		Phoneme &	Tf-dec-IMAB & 1.217{\tiny $\pm$0.007} & 1.032{\tiny $\pm$0.011} & \textbf{0.950{\tiny $\pm$0.011}}\\
		Spectra &	Tripletformer & \textbf{1.180{\tiny $\pm$0.008}} & \textbf{1.025{\tiny $\pm$0.005}} & {0.975{\tiny $\pm$0.009}} \\
		\bottomrule
		
	\end{tabular}
\end{table}

\subsection{Additional Experiments for deterministic interpolation}
\label{sec:det_interpol}

In Table~\ref{tab:apdx_det}, we present the results for deterministic interpolation on all the datasets for mTAN, HETVAE and Tripletformer. We observe that Tripletformer provides best performance in $24$ out of $30$ comparisons.

\begin{table*}
	\centering
	\scriptsize
	\caption{Results for deterministic interpolation on all datasets. Evaluation measure is Mean Squared Error, lower the best.}
	\label{tab:apdx_det}
	\begin{tabular}{l|lll||lll}
		\toprule
		tp. obs. & $10\%$  & $50\%$ & $90\%$ & $10\%$  & $50\%$ & $90\%$ \\
		\hline
		& \multicolumn{3}{c||}{random missing} & \multicolumn{3}{|c}{burst missing} \\
		\hline
		&	\multicolumn{6}{c}{Physionet} \\
		mTAN  &    0.530{\tiny $\pm$0.001}    &    0.377{\tiny $\pm$0.001}    &    0.368{\tiny $\pm$0.002}    &    0.646{\tiny $\pm$0.001}    &    0.569{\tiny $\pm$0.002}    &    0.392{\tiny $\pm$0.001}\\
		HETVAE    &    0.549{\tiny $\pm$0.001}    &    0.378{\tiny $\pm$0.001}    &    0.343{\tiny $\pm$0.002}    &    0.635{\tiny $\pm$0.002}    &    0.567{\tiny $\pm$0.002}    &    \textbf{0.331{\tiny $\pm$0.001}}\\
		Tripletformer     &    \textbf{0.525{\tiny $\pm$0.006}}   &    \textbf{0.372{\tiny $\pm$0.007}}   &    \textbf{0.331{\tiny $\pm$0.011}}   &    \textbf{0.627{\tiny $\pm$0.001}}   &    \textbf{0.554{\tiny $\pm$0.004}}   &    0.335{\tiny $\pm$0.003}\\
		\hline
		&	\multicolumn{6}{c}{MIMIC-III} \\
		
		mTAN  &    0.661{\tiny $\pm$0.001}    &    0.477{\tiny $\pm$0.001}    &    0.474{\tiny $\pm$0.001}    &    0.805{\tiny $\pm$0.001}    &    0.710{\tiny $\pm$0.001}    &    0.511{\tiny $\pm$0.001}\\
		HETVAE    &    0.676{\tiny $\pm$0.001}    &    0.472{\tiny $\pm$0.001}    &    0.430{\tiny $\pm$0.001}    &    0.804{\tiny $\pm$0.002}    &    0.727{\tiny $\pm$0.002}    &    \textbf{0.446{\tiny $\pm$0.000}}\\
		Tripletformer     &    \textbf{0.646{\tiny $\pm$0.023}}   &    \textbf{0.462{\tiny $\pm$0.017}}   &    \textbf{0.414{\tiny $\pm$0.008}}   &    \textbf{0.788{\tiny $\pm$0.002}}   &    \textbf{0.684{\tiny $\pm$0.001}}   &    0.448{\tiny $\pm$0.002} \\
		\hline
		& \multicolumn{6}{c}{Physionet2019} \\
		
		mTAN  &    \textbf{0.581{\tiny $\pm$0.002}}   &    0.419{\tiny $\pm$0.002}    &    0.415{\tiny $\pm$0.005}    &    0.737{\tiny $\pm$0.001}    &    0.541{\tiny $\pm$0.001}    &    0.461{\tiny $\pm$0.002} \\
		HETVAE    &    0.593{\tiny $\pm$0.000}    &    0.410{\tiny $\pm$0.001}    &    \textbf{0.350{\tiny $\pm$0.000}}   &    0.744{\tiny $\pm$0.005}    &    0.515{\tiny $\pm$0.000}    &    0.485{\tiny $\pm$0.002}\\
		Tripletformer     &    0.585{\tiny $\pm$0.001}    &    \textbf{0.392{\tiny $\pm$0.001}}   &    0.371{\tiny $\pm$0.003}    &    \textbf{0.591{\tiny $\pm$0.009}}   &    \textbf{0.392{\tiny $\pm$0.001}}   &    \textbf{0.370{\tiny $\pm$0.003}}\\
		\hline    
		&	\multicolumn{6}{c}{PenDigits} \\
		
		mTAN  &    0.783{\tiny $\pm$0.004}    &     0.453{\tiny $\pm$0.007}   &    0.335{\tiny $\pm$0.018}    &    0.743{\tiny $\pm$0.003}    &    0.466{\tiny $\pm$0.005}    &    0.335{\tiny $\pm$0.018} \\
		HETVAE    &    0.814{\tiny $\pm$0.001}    &    0.725{\tiny $\pm$0.019}    &    0.848{\tiny $\pm$0.011}    &    0.736{\tiny $\pm$0.026}    &    0.723{\tiny $\pm$0.001}    &    0.848{\tiny $\pm$0.011}\\
		Tripletformer     &    \textbf{0.695{\tiny $\pm$0.008}}    &    \textbf{0.326{\tiny $\pm$0.003}}   &    \textbf{0.215{\tiny $\pm$0.016}}    &    \textbf{0.644{\tiny $\pm$0.001}}   &   \textbf{0.408{\tiny $\pm$0.004}}    &    \textbf{0.215{\tiny $\pm$0.016}}\\
		\hline
		&	\multicolumn{6}{c}{PhonemeSpectra} \\
		mTAN  &    0.698{\tiny $\pm$0.005}    &    0.409{\tiny $\pm$0.006}    &    0.515{\tiny $\pm$0.016}    &    0.896{\tiny $\pm$0.001}    &    \textbf{0.782{\tiny $\pm$0.002}}   &    0.817{\tiny $\pm$0.011} \\
		HETVAE    &    0.785{\tiny$\pm$0.002} &    0.385{\tiny $\pm$0.010}    &    0.528{\tiny$\pm$0.004} &    0.887{\tiny $\pm$0.001}    &    0.815{\tiny $\pm$0.001}    &    \textbf{0.808{\tiny $\pm$0.002}}\\
		Tripletformer     &    \textbf{0.679{\tiny $\pm$0.011}}   &    \textbf{0.367{\tiny $\pm$0.035}}    &    \textbf{0.210{\tiny $\pm$0.009}}    &    \textbf{0.854{\tiny $\pm$0.001}}   &    0.808{\tiny $\pm$0.001}    &    0.796{\tiny $\pm$0.002}\\
		\bottomrule
	\end{tabular}
\end{table*}

\subsection{Discussions}


While Tripletformer offers probabilistic interpolation using parametric distributions, it is susceptible to performance degradation if the assumed distribution mismatches the observed data.

Additionally, Tripletformer focuses on providing marginal distributions rather than joint distributions. The focus on joint distributions can lead to variability in results when arbitrary time points are included. To address these limitations, we plan to develop a model that employs joint distributions, potentially enhancing prediction smoothness and reducing variability for random target times.


%

\section*{Conclusions}
\label{sec:concl}

In this work, we propose a novel model called Tripletformer for the problem of probabilistic interpolation in Irregularly sampled Time Series with missing values (IMTS). Our Tripleformer consists of a transformer like architecture, operates on a set of observations. We employ induced multi-head attention block in the encoder in order to learn the restricted attention mechanism, and to circumvent the problem of quadratic computational complexity of the canonical attention. We experimented on $5$ real and synthetic IMTS datasets, various conditioning ranges and $2$ different sampling techniques: random missing and burst missing. Our experimental results attest that the proposed Tripletformer provides better interpolations compared to the state-of-the-art model HETVAE.
\section*{Acknowledgments}

\begin{small}
	This work was supported by the Federal Ministry for Economic Affairs and Climate Action (BMWK), Germany, within the framework of the IIP-Ecosphere project (project number: 01MK20006D), and Funded by the Lower Saxony Ministry of Science and Culture under grant number ZN3492 within the Lower Saxony ``Vorab'' of the Volkswagen Foundation and supported by the Center for Digital Innovations (ZDIN).
\end{small}
\bibliography{bib}
\bibliographystyle{ieeetr}

\appendix

\section{Hyperparameters searched for the baseline models}

Hyperparameters searched for the competing models

\spara{GPR} Following~\cite{SM22} we use squared exponential kernel. We search learning rate from $\left\{0.1, 0..1, 0.001 \right\}$ and batch size from $\left\{32, 64, 128, 256\right\}$.

\spara{M-GPR} In M-GPR we used convolution of kernels as shown in~\cite{DP14}. Similar to GPR, we search learning rate from $\left\{0.1, 0..1, 0.001 \right\}$ and batch size from $\left\{32, 64, 128, 256\right\}$. 

\spara{ODE models:} We search the standard deviation over the range of $\left\{0.01, 0.2,0.4,0.6, 0.8, 1.0, 1.2, 1.4, 1.6, 1.8, 2.0\right\}$, select the number of GRU hidden units, latent dimension, nodes in the fully connected network for the ode function in both encoder and decoder from $\left\{20,32,64,128,256 \right\}$. Number of fully connected layers are searched in the range of $\left\{1,2,3\right\}$.

\spara{mTAN:} We search the standard deviation from $\left\{0.01, 0.2,0.4,0.6, 0.8, 1.0, 1.2, 1.4, 1.6, 1.8, 2.0\right\}$, number of attention heads from $\left\{1,2,4\right\}$, reference points from $\left\{8,16,32,64,128\right\}$, latent dimensions from $\left\{20,30,40,50\right\}$, generator layers from $\left\{25,50,100,150\right\}$, and reconstruction layers from $\left\{32,64,128,256\right\}$. 

\spara{HETVAE:} We search the same hyperparameter range mention in~\cite{SM22}. We set time embedding dimension to 128, search hidden nodes in the decoder from $\left\{16,32,64,128\right\}$, number reference points from $\left\{4,8,16,32\right\}$, latent dimension from $\left\{8,16,32,64,128\right\}$, width of the fully connected layers from $\left\{128,256,512\right\}$ and augmented learning objective from $\left\{1.0,5.0,10.0\right\}$.


\end{document}